\documentclass[]{jmlr}

\usepackage{longtable}

\usepackage{natbib}
\usepackage{multirow}
\usepackage{booktabs}
\usepackage{subcaption}

\usepackage{booktabs}
\usepackage[load-configurations=version-1]{siunitx} 

\theorembodyfont{\upshape}
\theoremheaderfont{\scshape}
\theorempostheader{:}
\theoremsep{\newline}

\jmlrvolume{}
\jmlryear{}
\jmlrworkshop{}

\title[SIPA: A Simple Framework for Efficient Networks]{SIPA: A Simple Framework for Efficient Networks}

 \author{\Name{Gihun Lee}\thanks{Equal contribution} \Email{opcrisis@kaist.ac.kr}\\
        \Name{Sangmin Bae}\footnotemark[1] \Email{bsmn0223@kaist.ac.kr}\\
        \Name{Jaehoon Oh} \Email{jhoon.oh@kaist.ac.kr}\\
        \Name{Se-Young Yun} \Email{yunseyoung@kaist.ac.kr}\\
\addr KAIST}

\begin{document}
\maketitle

\begin{abstract}
\thispagestyle{empty}
With the success of deep learning in various fields and the advent of numerous Internet of Things (IoT) devices, it is essential to lighten models suitable for low-power devices. In keeping with this trend, \emph{MicroNet Challenge}, which is the challenge to build efficient models from the view of both storage and computation, was hosted at NeurIPS 2019. To develop efficient models through this challenge, we propose a framework, coined as \emph{SIPA}, consisting of four stages: Searching, Improving, Pruning, and Accelerating. With the proposed framework, our team, \emph{OSI AI}, compressed $334\times$ the parameter storage and $357\times$ the math operation compared to WideResNet-28-10 and took 4th place in the CIFAR-100 track at MicroNet Challenge 2019 with the top 10\% highly efficient computation. Our source code is available from \href{https://github.com/Lee-Gihun/MicroNet_OSI-AI}{https://github.com/Lee-Gihun/MicroNet\_OSI-AI}.
\end{abstract}
\begin{keywords}
efficient neural network, compression, acceleration, MicroNet challenge
\end{keywords}

\section{Introduction}
\label{sec:intro}
With the ubiquity of not only mobile phones but also the Internet of Things (IoT) devices such as smartwatches, lightening deep neural networks has become unavoidable since the devices have limited resources. However, lightweight models with fewer neurons or shallower depth tend to have lower performance. Various model compression methods like neural architecture search, pruning, and quantization have been studied along each path to mitigate performance degradation. In line with this trend, \emph{MicroNet Challenege} was hosted at NeurIPS 2019, and the goal of this challenge was to build an efficient model with respect to the number of parameters and the number of computations, while the performance has equal to or greater than the proposed threshold accuracy. The two criteria are used to grade the efficiency of models in this challenge\footnote{https://micronet-challenge.github.io/scoring\_and\_submission.html}:

\begin{itemize}
\item \textbf{Parameter Storage}: the number of parameters of the model in inference time.
\vspace{-5pt}
\item \textbf{Math Operation}: the mean number of multiply-adds operations per example in inference time.
\end{itemize}

We participated in \emph{CIFAR-100}\,\citep{krizhevsky2009learning} track of which the threshold accuracy was 80\%. For scoring, the two criteria are normalized relative to WideResNet-28-10\,\citep{zagoruyko2016wide}, which stores 36.5M parameters and has 10.49B multiply-adds computations. The lower the score, the more efficient the model is.

In this paper, we propose a simple framework  \emph{SIPA} to build efficient networks. As the first step in this framework, we \emph{search} a block-based baseline model suitable for a particular dataset, and \emph{improve} the found baseline model by taking advantage of existing training methods, e.g., data augmentation, label smooth, and a learning rate scheduler. Next, we \emph{prune} the improved model. Finally, we \emph{accelerate} computation through sample-dependent adaptive paths, where the inference path varies according to the confidence of the decision. To this aim, we introduce  a novel loss function, coined as softsmoothing loss, for preventing over-confidence. Through the proposed framework, our final model achieved the final score of \emph{0.0058} and took  4th place in the challenge with the top 10\% highly efficient computation. Our key contributions can be summarized as follows:
\begin{itemize}
    \item We propose a simple 4-step framework  for efficient networks: Searching, Improving, Pruning, and Accelerating. This simple framework gives an easy and general strategy for developing lightweight neural networks.
    \item We test various existing performance-efficient training methods  and verify which techniques are compatible or not.
    \item We show that lightweight models can be more effectively compressed by analyzing the effects of pruning factors.
    \item We accelerate computation by making samples follow adaptive computational paths through which easy samples (i.e., highly confident samples) exit earlier. We provide a comprehensive study for a sample-dependent architecture design: exit position, exit module, and exit condition.
\end{itemize}

\section{SIPA: Searching, Improving, Pruning, and Accelerating}
\label{sec:approach}

\begin{figure}[ht]
    \centering
    \includegraphics[width=\textwidth]{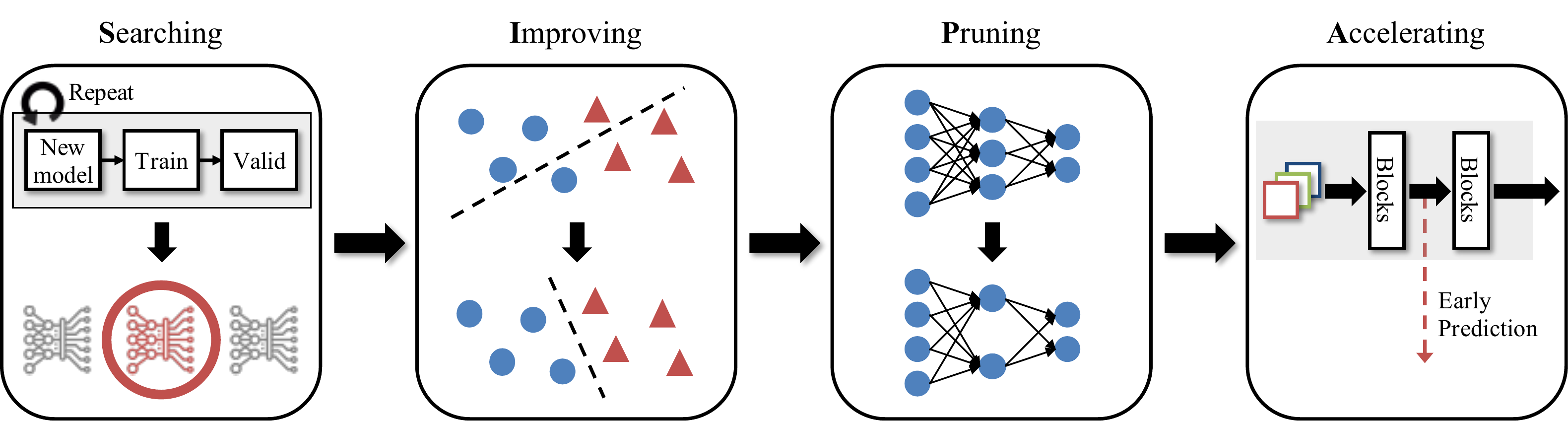}
    \caption{SIPA: A simple four-stage strategy for efficient networks.}
    \label{fig:strategy}
\end{figure}

For developing efficient networks, we establish a four-stage strategy depicted in \autoref{fig:strategy}. This strategy is simple yet effective with generality. This section is described in the order of the framework: Searching (Section~\ref{sec:Searching}), Improving (Section~\ref{sec:Improving}), Pruning (Section~\ref{sec:Pruning}), and Accelerating (Section~\ref{sec:Accelerating}).

\clearpage
\subsection{Searching}
\label{sec:Searching}
Searching is a step to find the baseline model structure through iterative validations, where the validation assesses each model by a fitness function based on performance and compactness, e.g., with positive constants $\alpha$ and $w$, \citep{dai2019chamnet} defines a fitness function of a vector of hyperparameters $x$ as follows: $R(x) = A(x) - [\alpha I (F(x, plat) - thres)]^w$, where $A(x)$, $I$, $F(x,plat)$, $plat$ and $thres$ are the accuracy, the unit step function, the latency of platform, target platform, and the resource constraint, respectively. However, searches often require a lot of computations since they are usually based on reinforcement learning \citep{zoph2016neural}. Therefore, it is important to restrict search spaces when sufficient resources are not allowed.

We use a popular block-based model  as an initial model to reduce the search cost.  With the block-based model, we search balanced coefficients of width, depth, and resolution using the compound scaling method proposed by \cite{DBLP:conf/icml/TanL19} to compensate any possible performance loss due to the reduced structure search space. In summary, we search in two steps.

\begin{itemize}
\item \textbf{Step 1 (Block arguments search)}: Most of recent convolutional networks exploit block-based architectures, e.g., WideResNet\,\citep{zagoruyko2016wide}, MobileNetV2\,\citep{sandler2018mobilenetv2}, and EfficientNet\,\citep{DBLP:conf/icml/TanL19}. These models are good initial models to reduce the search space greatly because the number and order of layers that compose a block are given. Therefore, we just need to search kernel size, stride, input channels, and output channels for each block.
\item \textbf{Step 2 (Scaling coefficients search)}: EfficientNet\,\citep{DBLP:conf/icml/TanL19} proposes the compound scaling rule, defined 
with a compound coefficient $\phi$ so that network width, depth, and resolution scale in a way: $\alpha^\phi$, $\beta^\phi$, and $\gamma^\phi$, respectively, with a constraint 
$\alpha \cdot \beta^2 \cdot \gamma^2 \approx 2$. Here, balancing among depth, width, and input resolution scaling coefficients (i.e., $\alpha$, $\beta$, and $\gamma$) is the key to improve performance under available resources. We thus search the three scaling coefficients.
\end{itemize}

Furthermore, Bayesian optimization \citep{shahriari2015taking} is incorporated for searching hyperparameters since  the Bayesian method could reduce the search cost through inference with prior knowledge.

\subsection{Improving}
\label{sec:Improving}

The baseline model obtained from the searching stage is expected to be at a desirable trade-off spot between model size and performance. In the improving stage, various methods are applied to improve the performance of the baseline model. These methods range from training settings(e.g., learning rate, batch size, and the number of epochs) to regularization or distillation. Any methods that can be applied to the training process are candidate methods. Through this stage, the model performance can be \emph{boosted} without or with negligible additional parameters.

\begin{figure}[h!]
    \centering
    \includegraphics[width=15cm]{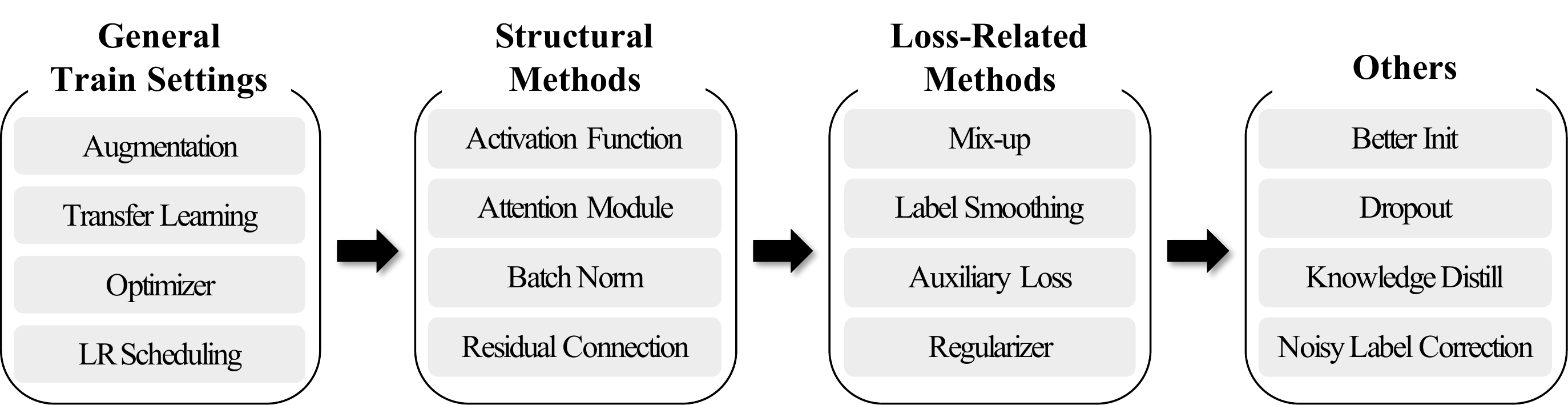}
    \vspace{-5pt}
    \caption{Procedure of improving stage.}
    \label{fig:improving_general}
\end{figure}

In this stage, we focus on how to obtain a desirable combination from a set of candidate techniques. It turns out that some methods are not compatible with others, and the effect of the method may vary on the efficient models that use much fewer parameters than general architectures. 
However, the number of possible combinations grows exponentially as $O({2}^{n})$ with respect to $n$, the number of candidate methods. Therefore, it is computationally intractable to examine all the combinations. To cope with the difficulty of having a wide variety of combinations, we use a greedy-based approach. More precisely, a method is added and adopted only when it makes performance improvement. Although this cannot guarantee optimal combination, a useful sub-optimal combination is achievable.
In this process, to determine the order of methods to be tested, we first divide the methods into four categories: general training settings, structural methods, loss-related methods, and others, and then sort the method group from the least sensitive to the most sensitive with respect to the change of settings. This tactic reduces the chance of a specific volatile method leads to rejecting potentially beneficial methods. \autoref{fig:improving_general} illustrates the general procedure of the approach mentioned above.

\vspace{-25pt}

\subsection{Pruning}
\label{sec:Pruning}
Pruning is a popular compression method to remove unnecessary connections between neurons without significant performance degradation.
However, most pruning methods have been tested on large networks \citep{liu2018rethinking,li2016pruning}, and it is not clear whether the pruning algorithm still performs well with the efficient model obtained from the previous step. Recent lightweight networks such as MobileNetV2 or EfficientNet uses much less parameters, from which more performance degradation is expected with the same pruning ratio compared to the general networks. Considering the less over-parameterized setting, we analyze the pruning methods in four aspects.

\paragraph{A. Weight pruning vs. Filter pruning} Pruning methods are divided into weight pruning\, \citep{lecun1990optimal} and filter pruning\, \citep{molchanov2016pruning} according to pruning targets. Although filter pruning relieves the burden of using a library and hardware for the sparse matrix multiplications \citep{li2016pruning}, the way to prune the whole filters might not be good with the lightweight networks. For instance, in depthwise filters, which apply a single convolutional filter per input channel, the importance of a single filter significantly increases compared to the conventional convolution layer.
\vspace{-5pt}
\paragraph{B. Global vs. Layer-wise} The necessity of parameters is decided by comparison either in a global or in a layer-wise perspective. 
From a global perspective, the pruning ratio of each layer is determined by the overall comparison between all parameters.
On the other hand, from a layer-wise perspective, all layers have the same sparsity as a result of the magnitude comparison within each layer.
However, when the importance of each layer is different, having the same sparsity over all layers can cause a performance degradation.

\vspace{-5pt}
\paragraph{C. One-shot pruning vs. Iterative pruning}  One-shot pruning eliminates the target percentage of parameters in one round, while iterative pruning eliminates them in multiple rounds. Iterative pruning is inclined to preserve a better performance with the same pruning ratio compared to the one-shot pruning but requires more computations to finish the pruning process in general.
\vspace{-5pt}

\paragraph{D. Fine-tuning vs. Re-initialization} Retraining step is essential to ensure that the pruned model reproduces the performance of the original model. There are two approaches for retraining: fine-tuning and re-initialization. 
In fine-tuning, the pruned model continues to learn from the checkpoint weight of the trained model. On the other hand, re-initialization retrains from scratch with re-initialized weights. Since it trains from the initial weights, re-initialization requires much more retraining cost than fine-tuning. Moreover, re-initialization cannot achieve comparable accuracy with fine-tuning on a difficult dataset \citep{liu2018rethinking}.


\subsection{Accelerating}
\label{sec:Accelerating}

\begin{figure}[h!]
    \centering
    \includegraphics[width=0.95\linewidth]{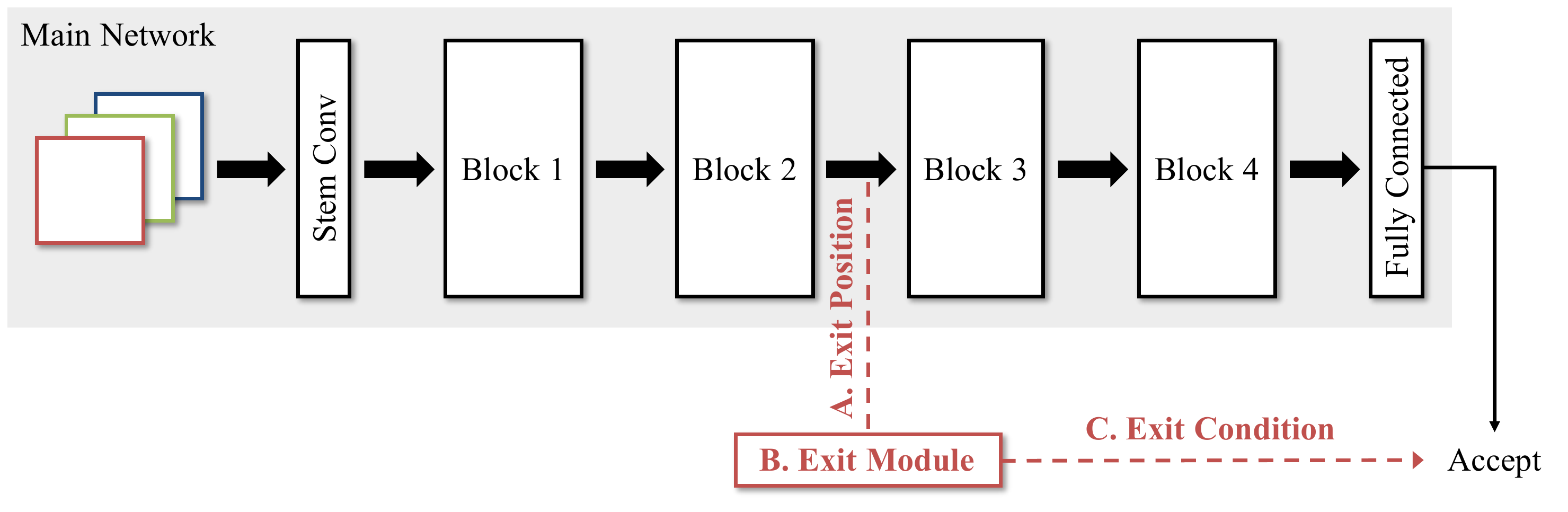}
    \caption{An overview of early prediction.}
    \label{fig:adaptive_comp}
\end{figure}

In this work, we introduce an exit model so that the neural network can predict easy samples earlier without changing the main network architecture. By using the sample-dependent adaptive paths, network inference time can be accelerated \citep{skipnet,scan}. 
A network model with a single early prediction path is illustrated in \autoref{fig:adaptive_comp}. Every sample is predicted first at the early prediction path after block 2. If the prediction result is confident enough, the prediction is accepted as the output and the inference process quits. Otherwise, further computation is conducted at later blocks, and the main network predicts the output.

\paragraph{A. Exit Position} 
As the exit position is located earlier, more computational costs are saved but less likely to obtain confident predictions since the features are not organized enough to be classified. Note that if the prediction at an exit position is not accepted, we spend extra computational cost at the exit position to check the prediction and the confidence. It means that selecting many exit positions might not help to accelerate inference.
\paragraph{B. Exit Module}
In the exit module, additional representation capacity is introduced to extract features because it is hard to predict directly from intermediate features, which mainly trained for prediction from the main path. The more capacity is added to the exit module, the better the performance of early prediction. However, the costs of the early prediction also increase with the bigger exit module.
\paragraph{C. Exit Condition}
Exit condition has to define how to measure the confidence and how to set the accept threshold. Here, we use the maximum value of softmax output as the confidence of prediction, but other confidence measures can be adopted as well.
The threshold setting controls the trade-off between the model accuracy and the computational cost. We very carefully tune the threshold to optimize the competition score.

To have a better confidence measure at the exit module, we propose a novel loss function, referred to as \emph{softsmoothing loss}. The loss function for the early prediction path ${L}_{ep}$ is defined as follows:
\begin{equation}
    {L}_{ep} = (1 + \textit{Confidence}(\hat{y})) \cdot H(\hat{y}, y)
    \label{eq:softsmoothing}
\end{equation}
where $H(\hat{y}, y)$ is cross-entropy loss between prediction probability $\hat{y}$ and target probability $y$. By giving more weights to samples with larger confidence in training batch, early prediction becomes more stable in the larger confidence region in the test time.

\section{Experiments}
\label{sec:experiments}
Our experiments through SIPA are detailed in this section. \autoref{tab:performance} shows the accuracy and scores of our model, which are normalized relative to WideResNet-28-10 (36.5M parameters and 10.49B FLOPS) and applied fake quantization\footnote{The details of scoring can be found at \href{https://github.com/Lee-Gihun/MicroNet_OSI-AI}{https://github.com/Lee-Gihun/MicroNet\_OSI-AI}.}. Here, $n$ and $s$ indicate the counted numbers and the normalized scores, respectively.

\begin{table}[ht]
\footnotesize
  \caption{MicroNet challenge scores per stage.}
  \label{tab:performance}
  \centering
  \begin{tabular}{c|c|cc|cc|c}
    \toprule
    Stage & Accuracy & Parameters($n$) & FLOPS($n$) & Parameters($s$) & FLOPS($s$) & Total\\
    \midrule
    Searching    & 73.47\% & 0.238M & 0.089B & 0.006534 & 0.008447 & 0.014981 \\
    Improving    & 80.47\% & 0.238M & 0.089B & 0.006534 & 0.008447 & 0.014981 \\
    Pruning      & 80.05\% & 0.103M & 0.034B & 0.002833 & 0.003267 & 0.006100 \\
    Accelerating & 80.04\% & 0.109M & 0.029B & 0.002995 & 0.002803 & 0.005798 \\
    \bottomrule
  \end{tabular}
\end{table}

\subsection{Searching}
Taking into account the challenge period and the computing power we have(three RTX Titan V GPU devices), we start with \emph{EfficientNet}\,\citep{DBLP:conf/icml/TanL19}, which is the most lightweight model among the top-level accuracy models, as an initial block-based model. The block of EfficientNet depicted in \autoref{fig:block} (Appendix \ref{sec:appendix_searching}) consists of a bottleneck residual block of MobileNetV2 \citep{sandler2018mobilenetv2} and a squeeze-and-excitation block of SENet\,\citep{hu2018squeeze}. Two sets of hyperparameters mentioned in Section \ref{sec:Searching} (i.e., block arguments and scaling coefficients) are searched through HyperOpt\,\citep{bergstra2013hyperopt}, which is a tool for Bayesian hyperparameter optimization, and a little manual tuning. The baseline model is given in \autoref{fig:main} (Appendix \ref{sec:appendix_searching}).

\subsection{Improving}
\begin{figure}[h!]
    \centering
    \includegraphics[width=15cm]{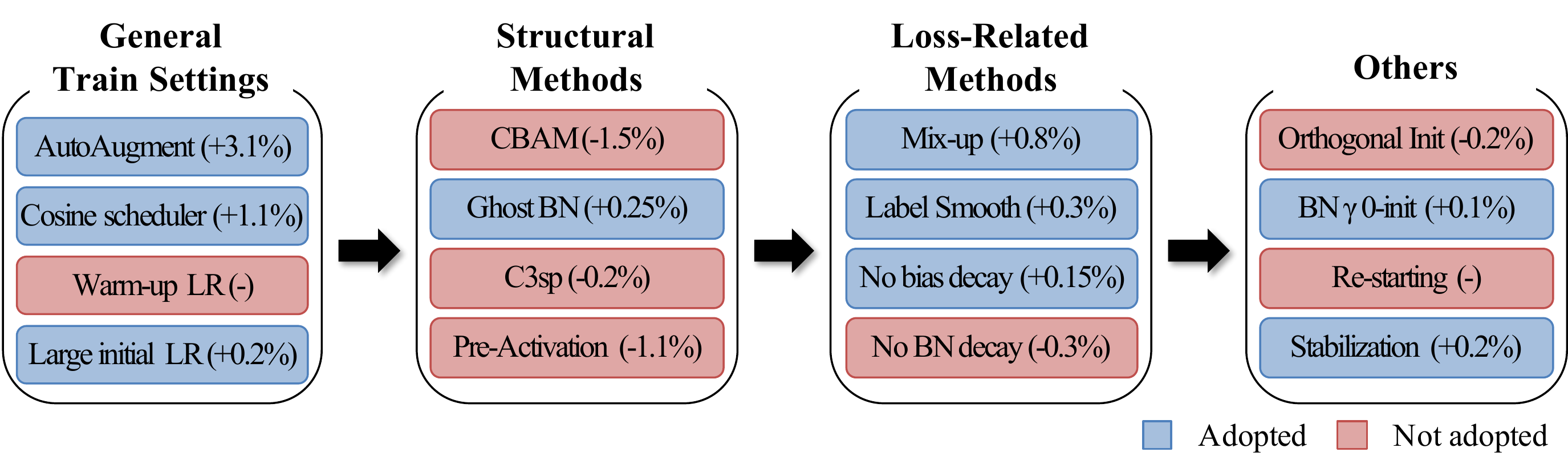}
    \vspace{-10pt}
    \caption{Order of applied improving methods in our approach.}
    \label{fig:improving_implementation}
\end{figure}

Our approach for the improving stage is outlined in \autoref{fig:improving_implementation}. The amount of improvement is estimated by comparing the average performance difference between groups with and without each method. Indeed, this stage increases the accuracy of our model significantly. The detailed descriptions of each method and how it applies to, including the method that does not work with our approach, is explained in Appendix \ref{sec:appendix_improving}.

\vspace{-5pt}
\vspace{-20pt}

\subsection{Pruning}
With our efficient network and limited computation resources, it is better to iteratively prune the smallest individual parameters from the global perspective and fine-tune from the checkpoint of the trained model. Furthermore, we propose a layer-wise normalization technique to balance the norm values of each layer and to prevent imbalanced pruning between layers. Because the lightweight model shows a more sensitive response to the change of pruning ratio, it is better to keep or decrease the pruning ratio per round for accuracy maintenance. Results of model accuracy, according to pruning ratio, are detailed in \autoref{fig:pruning} (Appendix \ref{sec:appendix_pruning}).
\vspace{-8pt}

\subsection{Accelerating}
Since our network is extremely small, selecting more than one path results in a bad trade-off between computational cost and accuracy. We search multiple positions using our simple exit module design depicted in \autoref{fig:arch} (Appendix \ref{sec:appendix_accelerating}). The detailed result of early prediction performance for exit positions is reported in Appendix \ref{sec:appendix_accelerating}. The parameters of the main network are frozen during early prediction path training to avoid hurting the main network performance. After training, a confidence threshold is decided using the validation set.
\vspace{-8pt}

\section{Discussion}
\paragraph{Rethinking Efficient Networks} 
We have several interesting observations that could give useful insights for designing efficient networks. In the searching stage, the expansion ratio, which controls how much the input channels are expanded, is the most sensible factor among block arguments. Moreover, increasing input resolution via nearest neighbor interpolation, where an upscaled point uses the same intensity of its nearest neighbor, improves performance considerably. The observations suggest that the performance bottleneck of efficient networks is on preserving the information of the previous step, and expanding the information to a high-dimensional space is prone to maintain essential information. We also observe that the pruned model sometimes shows better performance than before. This phenomenon implies that the unnecessary parameters of the network are more related to the noise in the inference of the model rather than the essential information to be exploited in inference. 
\vspace{-5pt}

\paragraph{Network Quantization}
It is found that quantizing the network parameters to 16-bit does not hurt performance at any stage with proper fine-tuning. We use 16-bit training from the beginning and there is only negligible performance degrade. Although more recent quantization methods  with 8-bit \citep{8bit-training} or 4-bit \citep{4bit-training} are not tested  in this paper, we recommend applying from the beginning with any quantization methods for the stability of training.

\begin{figure}[h]
    \centering
    \includegraphics[width=0.9\linewidth]{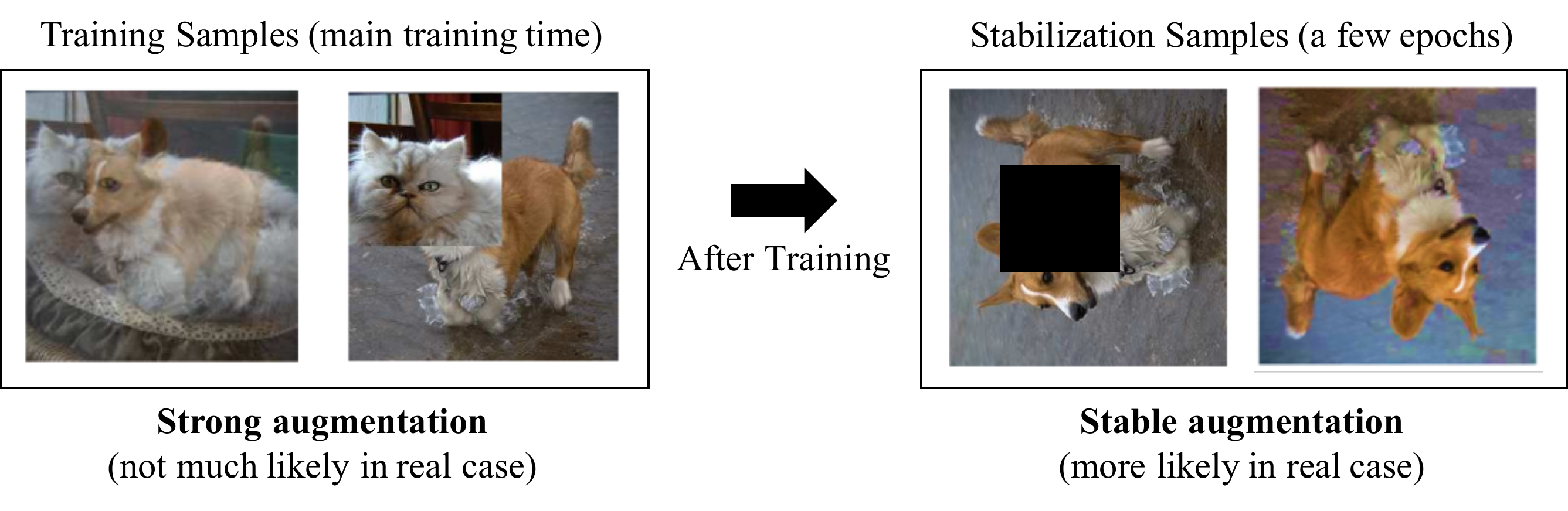}
    \vspace{-10pt}
    \caption{An overview of learning stabilization.}
    \label{fig:stabilization}
\end{figure}

\vspace{-5pt}
\paragraph{Distribution Stabilization}
Although strong data augmentation methods show impressive performance improvements, we also observe that too strong  augmentations, e.g., \citep{mixup, cutmix, autoaugment}, sometimes induce the model to learn with samples that are unlikely to be in the real world. To fix this problem, we suggest an additional fine-tuning step stabilize learned distribution. By training a few epochs with augmented samples, which are \emph{more likely in real-world}, the learned distribution can be shifted close to the test time distribution. \autoref{fig:stabilization} shows an overview of how our proposed stabilization step works. We find that training a few epochs with a small learning rate can stabilize the learned distribution, without hurting generalization ability.

\vspace{-10pt}
\vspace{-25pt}

\section{Conclusion}
\label{sec:conclusion}
\vspace{-5pt}
There is a growing interest in building efficient network models to reduce computational and storage costs. In this work, we introduce a simple yet effective framework to build an efficient network. This paper describes key components to build our model, which was submitted to NeurIPS 2019 MicroNet challenge and took 4th place. The framework consists of four stages: searching, improving, pruning, and accelerating. This framework is not limited to a specific model but can be applied to any models with general convolutional network architectures. We find that several improvement techniques are incompatible and even hurt performance in a lightweight network. While many pruning methods have been claimed to attain a 90-95\% pruning ratio in large models, pruning for the already small model is another story. We use iterative pruning and layer normalization to preserve model performance for the pruning process. In the accelerating step, a simple extension module  enables the early prediction for easy samples without changing the main architecture. We hope that our framework can enlarge the deployment of efficient network models.

\clearpage

\bibliography{references}

\clearpage
\appendix
\section{Details of the experiment results}
\label{sec:appendix_detail}
Appendix \autoref{sec:appendix_detail} contains details of the experimental results in the order of our framework. The found baseline model and EfficientNet block are described in Appendix \ref{sec:appendix_searching}, the improving techniques that tested in our improving stage is explained in Appendix \ref{sec:appendix_improving}. Appendix \ref{sec:appendix_pruning} shows the varying performance of the model by iterative pruning. Finally, the early prediction results and effect of softsmoothing loss is described in Appendix \ref{sec:appendix_accelerating}.

\subsection{Searching}
\label{sec:appendix_searching}
\begin{figure}[ht]
    \centering
    \includegraphics[width=\linewidth]{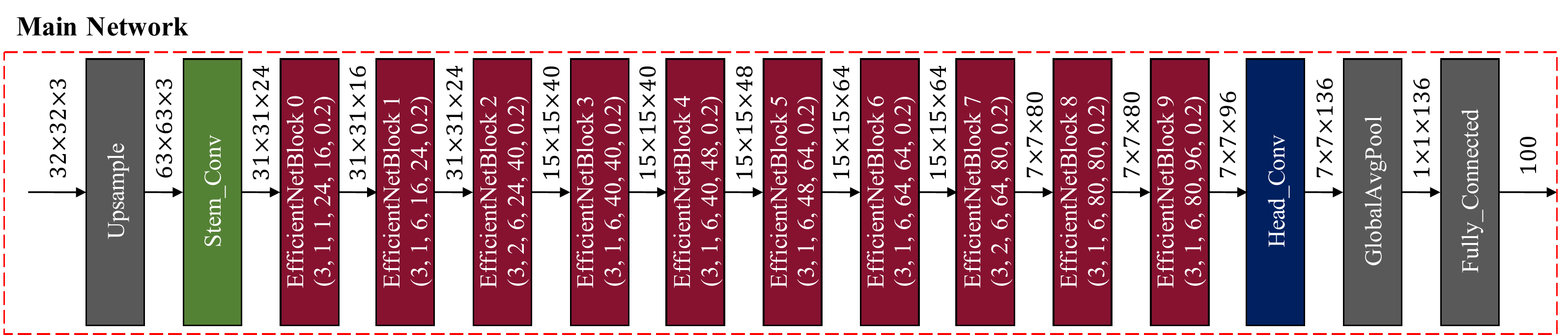}
    \begin{minipage}[t]{\linewidth}
        \centering
        \subcaption{Main network}\label{fig:main}
    \end{minipage}
    \includegraphics[width=\linewidth]{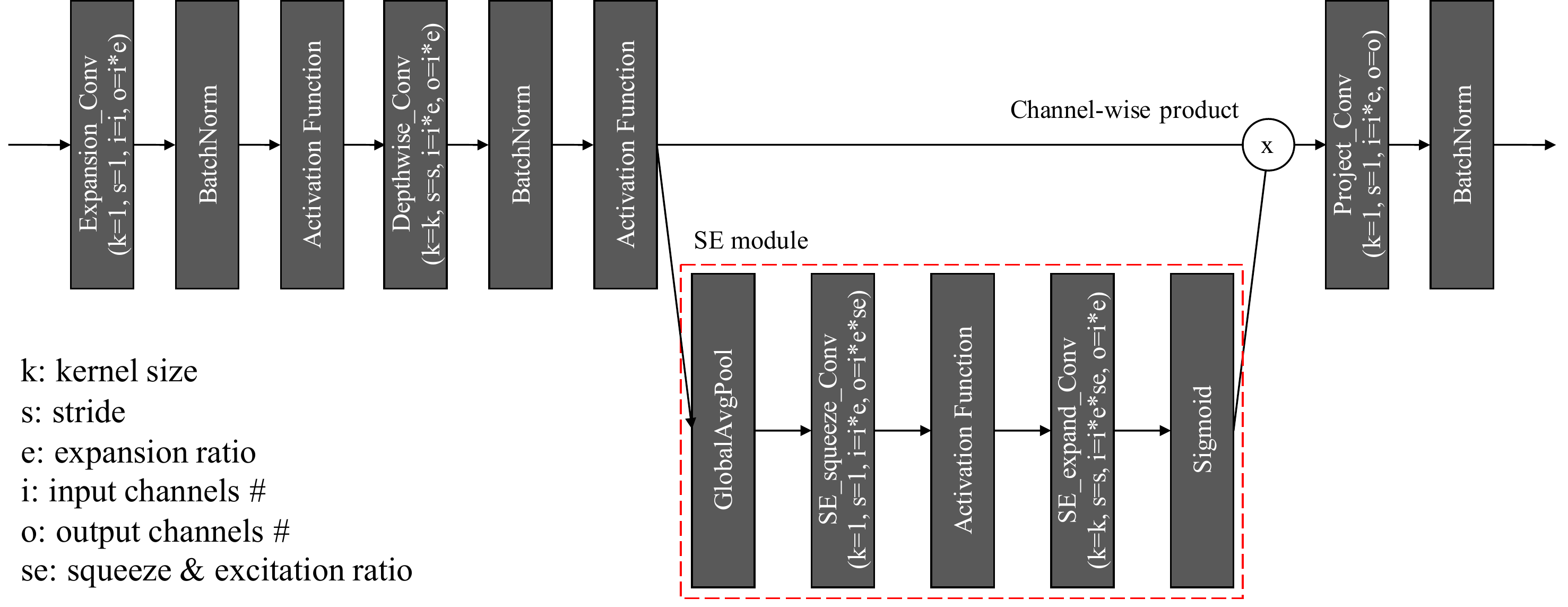}
    \begin{minipage}[t]{\linewidth}
        \centering
        \subcaption{EfficientNet block ($k$, $s$, $e$, $i$, $o$, $se$)}\label{fig:block}
    \end{minipage}
    \caption{Architecture overview.}  
\end{figure}

Through searching, block arguments of EfficientNet and three scaling coefficients, $\alpha=1$, $\beta=0.9$, and $\gamma=1.4$, are searched. \autoref{fig:main} describes our baseline model and \autoref{fig:block} describes an EfficientNet block.

\clearpage
\subsection{Improving}
\label{sec:appendix_improving}
\vspace{-3pt}
\paragraph{AutoAugment} searches dataset-specific augmentation policy based on a method using HyperOpt on the augmentation space proposed in \citep{autoaugment}. We search 25 sub-policies for the CIFAR-100 dataset.
\paragraph{Cosine annealing scheduler} adjusts the learning rate as a cosine function \citep{cosine_scheduler}. The learning rate ${\eta}_{t}$ is defined as ${\eta}_{t} = {\eta}_{min} + \frac{1}{2}({\eta}_{max}-{\eta}_{min})(1+cos (\frac{{T}_{cur}}{{T}_{max}}\pi))$, where ${\eta}_{min}$ and ${\eta}_{max}$ are minimum and maximum learning rate. ${T}_{cur}$ is the number of epochs elapsed, and ${T}_{max}$ is the maximum number of epochs. Although the original method includes restarting and warm-up, we can not see the performance difference in our work and just apply as a monotonically decreasing learning rate scheduler.
\paragraph{Increasing initial learning rate} is known to help the model learned by the SGD algorithm converge to flat minima, which tends to show better generalization \citep{understandingBN}. We use an initial learning rate of 0.13, which is larger than 0.1 in standard settings.
\paragraph{No bias decay \& No Batch Normalization (BN) decay} help generalization, which is not using weight decay to the bias parameters and batch normalization parameters, respectively. Because these parameters are little portion, compare to the whole parameters, decaying these parameters does not have a significant effect on regularization. Rather, it results in under-fitting problem \citep{bag_of_tricks}. In our case, only not applying weight decay to bias works, while the other does not.
\paragraph{Ghost batch normalization} calculates BN statistics ($\beta$ and $\gamma$) from small virtual batch splits \citep{hoffer2017train}. This has a similar effect of training with small batch size while maintaining the training speed of when using a large batch size. We split batch size 128 into 4 virtual batches with size 32.
\paragraph{Mix-up training} uses training samples that consist of a convex combination of two different classes, where the combining coefficient is sampled from beta distribution \citep{mixup}. This results in the model can learn more smooth distribution and relieves over-fitting because there is a scarce chance to generate the same sample.
\paragraph{Label smoothing} softens target distribution by giving 1-$\epsilon$ to the correct class and disperses $\epsilon$ to the other classes \citep{szegedy2016rethinking}. Although the exact behavior of this method is not discovered, it still is a widely used method. We used $\epsilon$ as 0.3.
\paragraph{Activation function} adds non-linearity to models. We examine several activation functions such as CeLU, ReLU, Swish, and leaky-ReLU and find Swish activation generalizes well with a relatively less computational cost. Swish activation function \citep{ramachandran2017searching}  is $f(x) = x \cdot \text{sigmoid}(\beta x)$ where $\beta$ is a hyperparameter and we use $\beta$ as 1.
\paragraph{No validation data} is not splitting dataset for validation. After optimizing all the hyperparameters using the validation dataset, we use both the training dataset and the validation dataset for the final training.
\paragraph{Others} There are some techniques that we test but do not improve performance in our cases, such as refurbishing mis-labeled data in CIFAR-100 \citep{selfie} and symmetric padding \citep{c2sp}.

\clearpage
\subsection{Pruning}
\label{sec:appendix_pruning}
\begin{figure}[ht]
    \begin{minipage}[t]{\linewidth}
        \footnotesize
        \centering
          \begin{tabular}{c|c|c|c}
            \toprule
            Round & Pruning Ratio per Round & Sparsity & Accuracy\\
            \midrule
            0 & 0\% & 0\%(baseline) & 80.47\% \\ \midrule
            1 & 10\% & 10\% & 80.41\% \\
            2 & 10\% & 20\% & 80.30\% \\
            3 & 10\% & 30\% & 80.31\% \\
            4 & 10\% & 40\% & 80.02\% \\
            5 & 10\% & 50\% & 80.08\% \\ \midrule
            6 & 2.5\% & 52.5\% & 80.11\% \\
            7 & 2.5\% & 55\% & 80.09\% \\
            8 & 2.5\% & 57.5\% & 80.21\% \\
            9 & 2.5\% & 60\% & 80.36\% \\ \midrule
            10 & 2\% & 62\% & 79.89\% \\
            \textbf{11} & \textbf{2\%} & \textbf{64\%} & \textbf{80.05\%} \\
            12 & 2\% & 66\% & 79.57\% \\
            13 & 2\% & 68\% & 79.21\% \\
            14 & 2\% & 70\% & 79.19\% \\
            \bottomrule
          \end{tabular}
    \end{minipage}
    \centering
    \includegraphics[width=0.6\linewidth]{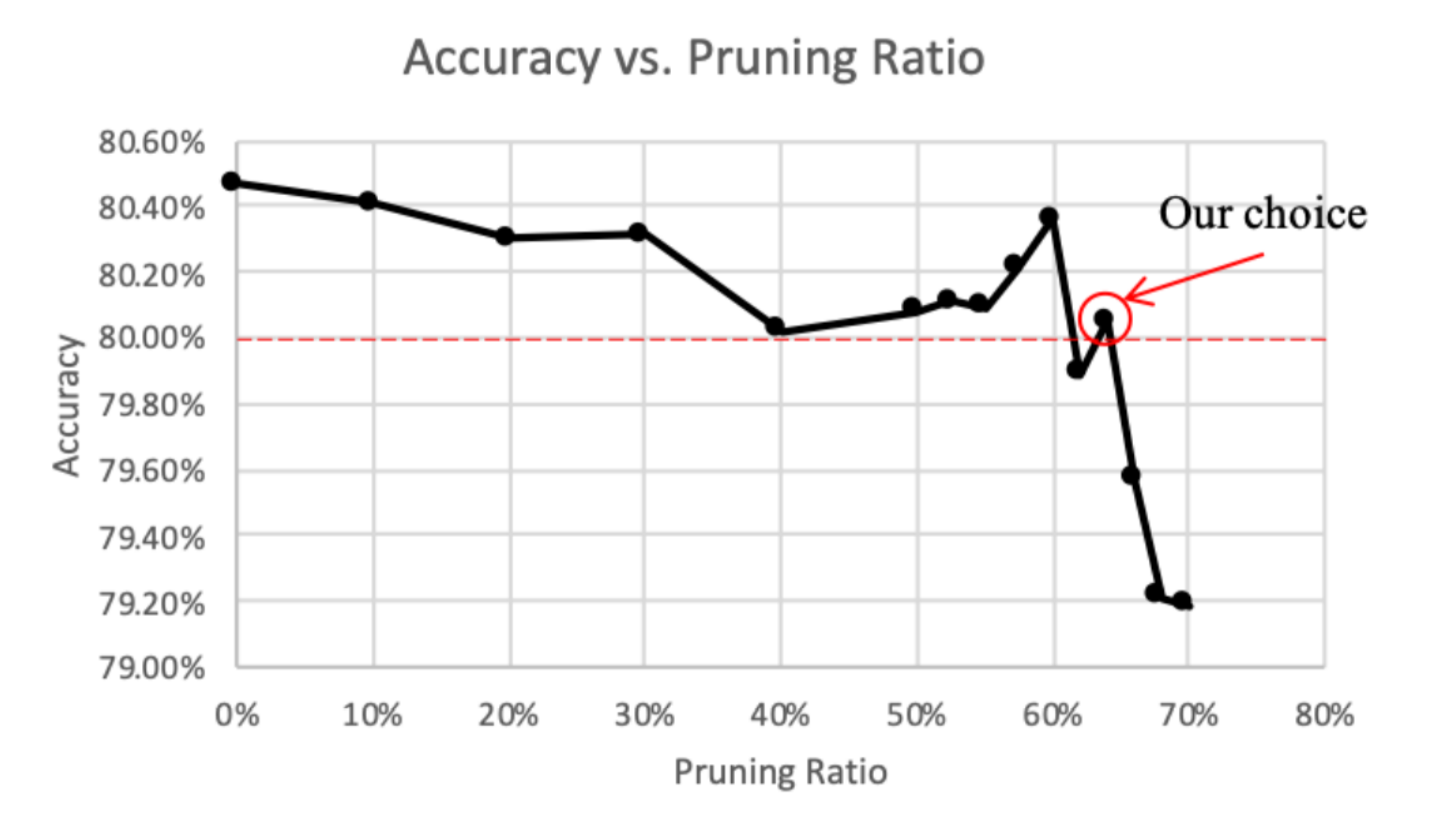}
    \begin{minipage}[t]{\linewidth}
        \centering
        \vspace{-20pt}
    \end{minipage}
    \vspace{-15pt}
    \caption{Results of iterative pruning.}
    \label{fig:pruning} 
\end{figure}

\autoref{fig:pruning} describes our pruning results per round. The bold line indicates the round of iterative pruning, which is selected as the result of the pruning stage. We empirically observe that significant performance reduction begins when more than 50\% of parameters are pruned. After reaching 50\% sparsity, we control the pruning ratio per round.

\vspace{-20pt}

\subsection{Accelerating}
\label{sec:appendix_accelerating}

\begin{figure}[h!]
    \centering
    \includegraphics[width=\linewidth]{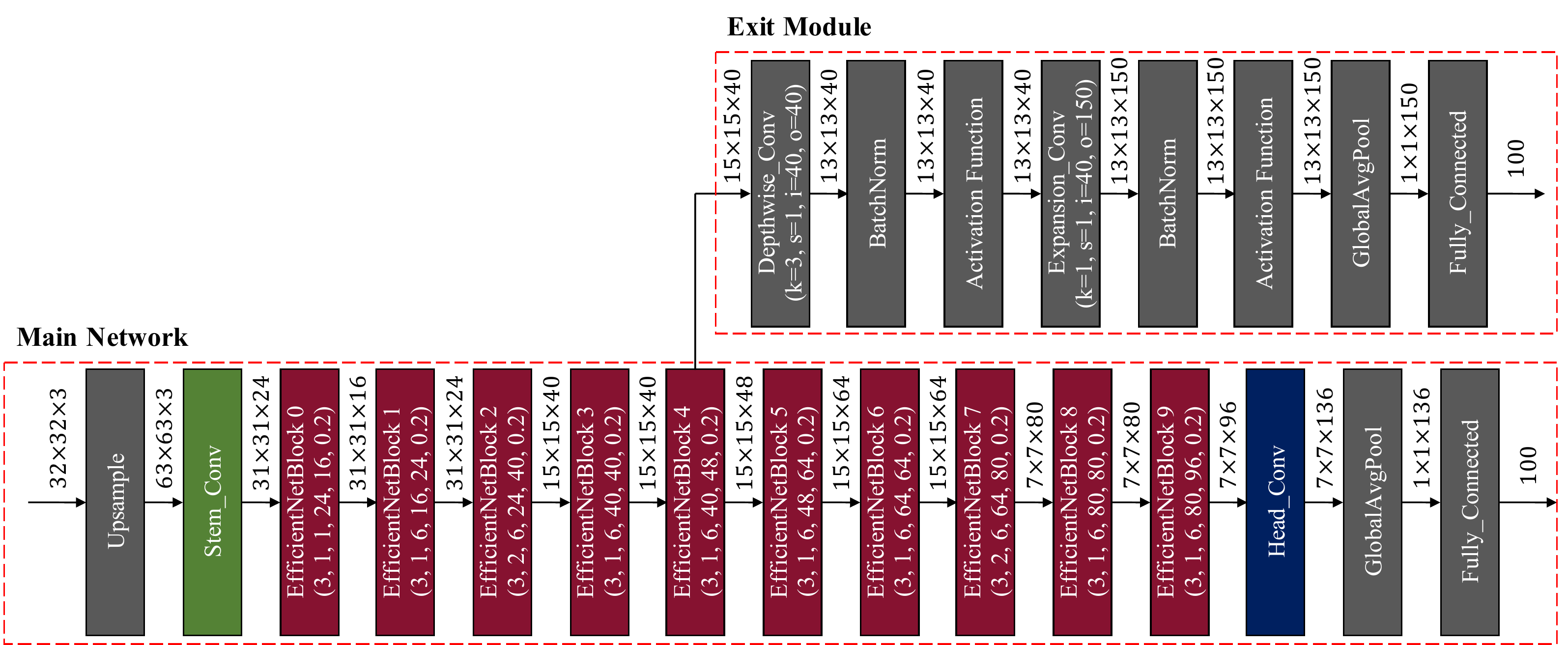}
    \caption{Main network \& exit module.}
    \label{fig:arch}
\end{figure}

\begin{figure}[h!]
    \centering
    \includegraphics[width=\linewidth]{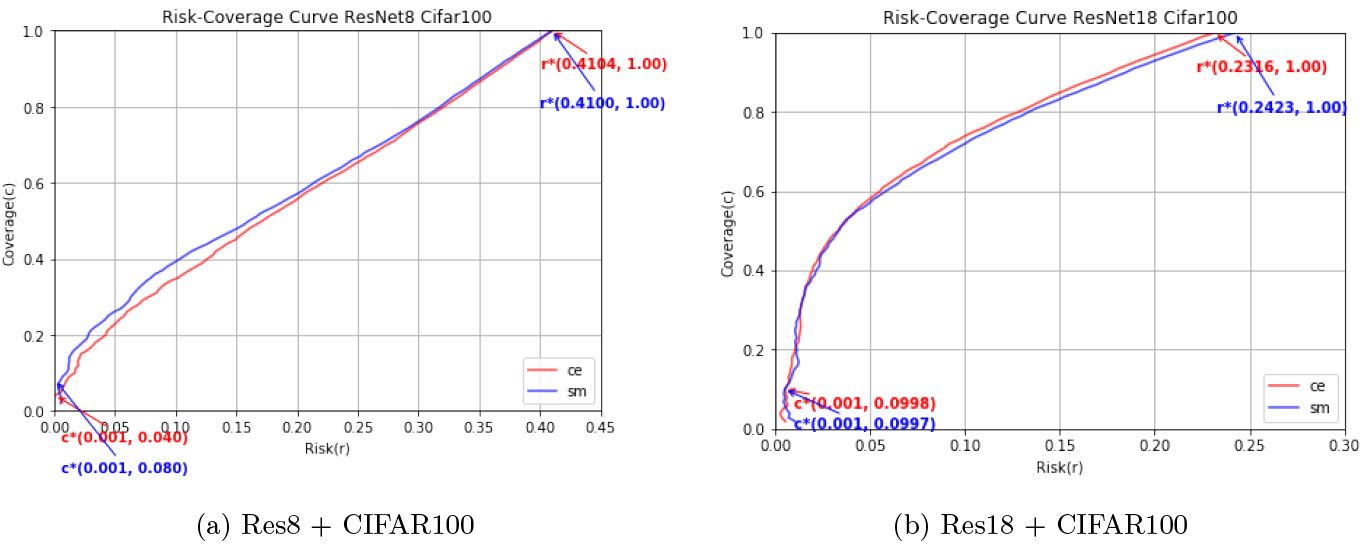}
    \caption{Effect of softsmoothing loss.}
    \label{fig:softsmoothing}
\end{figure}

 \autoref{tab:early_pred} shows the results according to the exit position. The Path FLOPs and Added Params show the cost ratio of early prediction path to the main path when exit from the corresponding Exit Position. Exit Ratio and Total Acc is tested with a fixed candidate confidence threshold (0.85, 0.88, 0.92). Exit position 4 with confidence threshold 0.85 is selected as the result of the accelerating stage. By further pruning and stabilizing for the early prediction path, we acquire our final model with 80.04\% accuracy. From these results, the overall structure of our model with an early prediction path is illustrated in \autoref{fig:arch}. Besides, \autoref{fig:softsmoothing} shows the effect of softsoomothing loss on a risk-coverage curve \citep{selective}. Softsmoothing loss(sm) shows low risk in high coverage region compare to cross-entropy loss(ce).
 
\begin{table}[h!]
\caption{Early prediction results for different exit positions.}
\label{tab:early_pred}
\centering
\scriptsize
\begin{tabular}{|c|c|c|c|c|c|c|}
\hline
Exit Position            & Path Acc                 & Path FLOPS               & Added Params            & Threshold    & Exit Ratio   & Total Acc\\ \hline
\multirow{3}{*}{Block 2} & \multirow{3}{*}{55.81\%} & \multirow{3}{*}{21.19\%} & \multirow{3}{*}{9.53\%} & 0.85         & 20.43\%      & 79.94\%  \\ \cline{5-7} 
                         &                          &                          &                         & 0.88         & 18.05\%      & 80.03\%  \\ \cline{5-7} 
                         &                          &                          &                         & 0.92         & 16.25\%      & 80.09\%  \\ \hline
\multirow{3}{*}{Block 3} & \multirow{3}{*}{58.66\%} & \multirow{3}{*}{29.58\%} & \multirow{3}{*}{9.53\%} & 0.85         & 22.92\%      & 80.26\%  \\ \cline{5-7} 
                         &                          &                          &                         & 0.88         & 20.24\%      & 80.35\%  \\ \cline{5-7} 
                         &                          &                          &                         & 0.92         & 18.33\%      & 80.38\%  \\ \hline
\multirow{3}{*}{\textbf{Block 4}} & \multirow{3}{*}{\textbf{64.62\%}} & \multirow{3}{*}{\textbf{38.91\%}} & \multirow{3}{*}{\textbf{10.10\%}} & \textbf{0.85} & \textbf{36.31\%}      & \textbf{79.94\%}  \\ \cline{5-7} 
                         &                          &                          &                         & 0.88         & 33.27\%      & 80.11\%  \\ \cline{5-7} 
                         &                          &                          &                         & 0.92         & 31.11\%      & 80.19\%  \\ \hline
\multirow{3}{*}{Block 5} & \multirow{3}{*}{69.85\%} & \multirow{3}{*}{52.72\%} & \multirow{3}{*}{11.23\%}& 0.85         & 50.42\%      & 79.96\%  \\ \cline{5-7} 
                         &                          &                          &                         & 0.88         & 47.05\%      & 80.00\%  \\ \cline{5-7} 
                         &                          &                          &                         & 0.92         & 44.92\%      & 80.14\%  \\ \hline
\end{tabular}
\end{table}

\vspace{-20pt}

\end{document}